\documentclass[conference]{IEEEtran}

\usepackage{cite}
\usepackage{amsmath,amssymb,amsfonts}
\usepackage{algorithmic}
\usepackage{graphicx}
\usepackage{textcomp}
\usepackage{xcolor}
\usepackage{authblk}
\usepackage{comment}
\usepackage{hyperref}
\usepackage{subcaption}
\usepackage{dblfloatfix}

\def\BibTeX{{\rm B\kern-.05em{\sc i\kern-.025em b}\kern-.08em
    T\kern-.1667em\lower.7ex\hbox{E}\kern-.125emX}}

\begin{document}

\title{Cross-Dataset Experimental Study of Radar-Camera~Fusion in Bird's-Eye View}

\author[1,2]{Lukas~St\"acker}
\author[1]{Philipp~Heidenreich}
\author[3]{Jason~Rambach} 
\author[2,3]{Didier~Stricker} 
\affil[1]{Stellantis, Opel Automobile GmbH, R\"usselsheim am Main, Germany}
\affil[2]{Rheinland-Pf\"alzische Technische Universit\"at Kaiserslautern-Landau, Kaiserslautern, Germany}
\affil[3]{German Research Center for Artificial Intelligence, Kaiserslautern, Germany}
\affil[ ]{\tt \normalsize lukas.staecker@external.stellantis.com}

\maketitle

\begin{abstract}
By exploiting complementary sensor information, radar and camera fusion systems have the potential to provide a highly robust and reliable perception system for advanced driver assistance systems and automated driving functions. Recent advances in camera-based object detection offer new radar-camera fusion possibilities with bird's eye view feature maps. In this work, we propose a novel and flexible fusion network and evaluate its performance on two datasets: nuScenes and \mbox{View-of-Delft}. Our experiments reveal that while the camera branch needs large and diverse training data, the radar branch benefits more from a high-performance radar. Using transfer learning, we improve the camera's performance on the smaller dataset. Our results further demonstrate that the radar-camera fusion approach significantly outperforms the camera-only and radar-only baselines.
\end{abstract}


\section{Introduction}

Automotive radar technology has become an important building block for advanced driver assistance systems and automated driving functions, where the goal is to make driving safer and more comfortable. Radar has several advantages when compared to LiDAR or camera: it is more robust and preferred in adverse weather conditions, it provides a higher range and direct measurements of the relative radial velocity, and often it is more affordable than LiDAR. 
In particular, radar-camera fusion currently appears to be the most relevant sensor combination in the automotive industry, providing complementary environmental information: radar sensors provide accurate distance and velocity measurements, while cameras provide accurate angular and rich semantic information.

Machine learning methods for 3D object detection have advanced significantly in the last years with the availability of large annotated datasets. These are traditionally dominated by LiDAR and camera perception, due to the fact that automotive radar used to have limited performance in terms of angular resolution, point cloud density, and object classification capabilities. However, these limitations are gradually dissolving with the development towards high-performance radars \cite{EngelsEtal2021} and the emergence of corresponding datasets \cite{ZhouEtal2022}. In this work, we select two suitable radar datasets for our experimental study: the nuScenes \cite{nuScenes2020} and the \mbox{View-of-Delft} \cite{ViewOfDelft2022} dataset. 

A recent trend in 3D object detection is to transform camera features into a common bird's eye view (BEV) representation, which offers a flexible architecture for fusion, either among multiple cameras or with a ranging sensor. 
In this work, we extend the BEVFusion \cite{BEVFusion2023} method, that was originally developed for LiDAR-camera fusion, to perform radar-camera fusion.
We train and evaluate our presented fusion method with the selected radar datasets. In several experiments, we discuss strengths and weaknesses of each dataset. Finally, we apply transfer learning to achieve further improvements.

\subsection{Radar Datasets}

Over the last few years, a large number of different automotive radar datasets for machine learning has been released. 
A recent overview is presented in \cite{ZhouEtal2022} 
\footnote{The corresponding author of \cite{ZhouEtal2022} regularly updates a curated list of radar datasets for detection, tracking and fusion methods on his \href{https://github.com/ZHOUYI1023/awesome-radar-perception}{github}.}, which includes traditional radar with a limited angular resolution, high-performance radars with denser point clouds, and non-automotive type scanning radar.
Note that traditional radars provide a 2+1D point cloud by measuring the range and azimuth angle, and the relative radial velocity in addition, whereas high-performance radars also measure the elevation angle and provide a 3+1D point cloud. 
There are datasets with different measurement data such as radar point clouds or pre-CFAR radar data, and different annotations such as 3D bounding boxes, semantic point cloud segmentation, or localization information. In this work, we focus on datasets with radar point clouds and 3D bounding boxes. 
Among the traditional radar datasets, we find the widely used nuScenes~\cite{nuScenes2020} dataset,  RadarScenes~\cite{RadarScenes2021} for semantic point cloud segmentation, and the recent aiMotive~\cite{aiMotive2022} dataset with focus on long range. 
Among the high-performance radar datasets, we find the small Astyx~\cite{Astyx2019} dataset, the \mbox{View-of-Delft}~\cite{ViewOfDelft2022} dataset with focus on vulnerable road users, and very recently TJ4DRadSet~\cite{TJ4DRadSet2022}, and K-Radar~\cite{KRadar2022}. 

For our experimental study, a reasonable number of annotated frames and objects is necessary. After careful examination of available datasets, we have selected the nuScenes and \mbox{View-of-Delft} dataset as suitable. For comparability, only the front-view of nuScenes is evaluated and only classes pedestrian, cyclist, and car are considered.

The nuScenes dataset covers 40k annotated frames, sampled at 2\,Hz. Each frame has measurements from six cameras with $70^\circ$ field of view (FoV), one Velodyne \mbox{HDL-32E} LiDAR, and five Continental ARS408 2+1D radars, for which only a limited number of radar detections is reported. To this end, multiple radar sweeps are accumulated to increase the point cloud density. 
The \mbox{View-of-Delft} dataset comes with 8.7k annotated frames, sampled at 10\,Hz. Each frame has measurements from one frontal camera with $64^\circ$ FoV, one Velodyne \mbox{HDL-64E} LiDAR, and one ZF \mbox{FR-Gen21} 3+1D radar, which provides a dense point cloud. For both datasets, the 3D bounding box annotations have been obtained with the help of the respective reference LiDAR.

\subsection{Related Work}

Approaches for fusion of cameras and ranging sensors such as radar or LiDAR have to solve the problem that the sensor data is available in different representations. 
While cameras provide images on the perspective plane, ranging sensors provide point clouds in 3D or BEV, making it difficult to associate features from the different modalities. 
Some approaches project the point cloud to the image to augment it with depth information \cite{Chadwick2019}. However, this projection is geometrically lossy, since only a small part of the image can be augmented. 
Other approaches do a reverse projection and decorate the point cloud with features extracted from the image \cite{Fei.2020}. Especially for sparse radar point clouds, this approach is semantically lossy. 
An alternative is to take object proposals from the camera and associate matching points from the point cloud to refine the proposals \cite{CenterFusion2021}. However, this limits the potential of the ranging sensor to a refinement, instead of enabling it to detect objects missed by the camera.

Recent advances in camera-only 3D object detection transform features from the camera into the BEV plane, enabling new possibilities for fusion. A recent technique to obtain BEV features from multiple cameras is the idea from Lift, Splat, Shoot \cite{LiftSplatShoot2020}. Here, a depth distribution for each pixel is unprojected into a frustum of features in 3D space, which is subsequently condensed into a rasterized BEV grid. This technique has been improved and extended for 3D object detection in \cite{BEVDet2021}. 
Based on this work, BEVFusion \cite{BEVFusion2023} has been proposed as a network for LiDAR-camera fusion. Here, voxelization and sparse 3D convolutions are used for LiDAR feature extraction before reducing the dimension to BEV. In our work, we further extend BEVFusion 
by replacing the LiDAR with radar and adapting a pillar feature encoding, similar to the one in \cite{PointPillars2019}.

\section{Radar-Camera Fusion in BEV}

\begin{figure}[ht]
\centerline{\includegraphics[width=\linewidth]{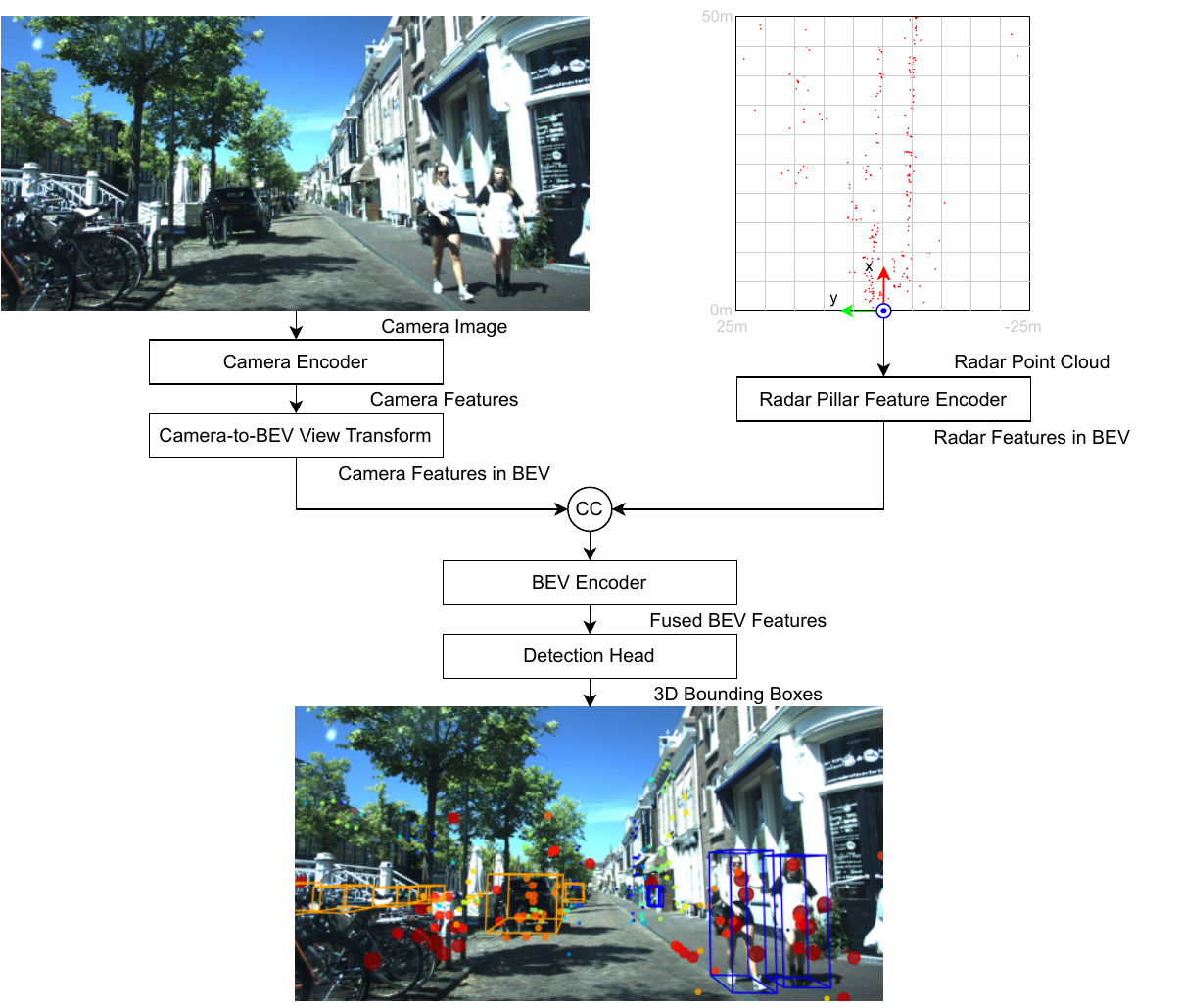}}
\caption{Flowgraph of the presented radar-camera fusion in BEV, based on \cite{BEVFusion2023}. In the resulting camera image, we include projected radar detections and ground truth bounding boxes.}
\label{BEVFusion}
\end{figure}

We follow the fusion architecture of \cite{BEVFusion2023}. An overview of our network for radar-camera fusion in BEV is provided in Figure~\ref{BEVFusion}. Note that the fusion occurs when the camera and radar features in BEV are concatenated. Below, we provide further detail for each block.

\subsection{Camera Encoder and Camera-to-BEV View Transform}

The camera encoder and view transform follow the idea of \cite{BEVDet2021}, which is a flexible framework to extract camera BEV features with arbitrary camera extrinsic and intrinsic parameters. First, the tiny version of a Swin Transformer \cite{SwinTransform2021} network is used  
to extract features from each image. Next, we transform the features from the image to the BEV plane using the Lift and Splat step from \cite{LiftSplatShoot2020}. To this end, a dense depth prediction is followed by a rule-based block, in which the features are transformed into a pseudo point cloud and rasterized and accumulated into the BEV grid.

\subsection{Radar Pillar Feature Encoder}

This block aims to encode the radar point cloud into BEV features on the same grid as the camera BEV features. To this end, we use the pillar feature encoding technique from \cite{PointPillars2019}, in which the point cloud is rasterized into voxels with infinite height, so-called pillars. 
Here, each point of the radar point cloud is represented by a position $(x, y, z)$, a radar cross section $\mathrm{RCS}$, a relative radial velocity $v_\mathrm{r}$, compensated by the ego motion, and a timestamp $t$. 
To encode geometrical relationships within voxels, we augment each point by the distance to the arithmetic mean of all points in the pillar $(x_\mathrm{c}, y_\mathrm{c}, z_\mathrm{c})$, and the distance to the pillar center ($x_\mathrm{p}, y_\mathrm{p}$), resulting in a feature vector of size 11. 
Note that for nuScenes, the radars are only 2+1D, so that there is no $z$-component and the feature vector is of size $9$.
All non-empty pillars are now processed using a simplified PointNet, before the resulting pillar features are populated back to the BEV grid.

\subsection{BEV Encoder}

Similar to \cite{BEVFusion2023}, the radar and camera BEV feature are fused by concatenation. The fused features are then processed by a joint convolutional BEV encoder to enable the network to account for spatial misalignment and use synergies between the different modalities.

\subsection{Detection Head}

We use the CenterPoint \cite{CenterPoint2021} detection head to predict heatmaps of object centers for each class. 
Further regression heads predict the dimension, rotation, and height of objects, as well as the velocity and class attribute for nuScenes. 
Whereas the heatmaps is trained with Gaussian focal loss, the remaining detection heads are trained with L1 loss.

\section{Experiments}

We use our presented radar-camera fusion in BEV to conduct several experiments on nuScenes and \mbox{View-of-Delft}. For a better comparability between the used datasets, we only use the frontal camera and the front-facing radars in nuScenes and filter out annotated 3D bounding boxes that fall outside the FoV of the frontal camera. For both datasets, we use a common BEV grid on $x\in[0,51.2]$ and $y\in[-25.6,25.6]$ with a step size of 0.1\,m. 
We further reduce the considered classes to pedestrian, cyclist and car, since these are the only classes with sufficient annotations in \mbox{View-of-Delft}. To account for potential class imbalances, we use class-balanced grouping and sampling \cite{CGBS2019}.
For both datasets, we aggregate the radar point cloud over 5 sweeps, corresponding to roughly 0.3\,s.

In addition to the presented radar-camera fusion, we use our object detection network also in radar-only and camera-only configurations, where we remove the encoded camera and radar feature inputs, respectively.

Since \mbox{View-of-Delft} has less camera frames and shows a smaller visual variability, we also investigate transfer learning of the camera feature encoding. To this end, we pretrain the camera-only network with nuScenes and fine-tune the camera-only and radar-camera fusion networks with \mbox{View-of-Delft} to improve the performance.

\subsection{Training and Evaluation}

We train and evaluate our networks with the official training and validation splits of nuScenes and \mbox{View-of-Delft}. 
We use an AdamW optimizer with a learning rate of 2e-4 for batch size 32 and a cyclic learning rate policy. 
Due to the different dataset sizes, we train nuScenes for 20 epochs and \mbox{View-of-Delft} for 80 epochs, and select the best-performing model on the validation set. 

To evaluate the 3D object detection performance, we consider the commonly used average precision (AP) and mean AP (mAP) metrics, where the latter corresponds to an average over relevant classes. 
The AP is defined as the area under the precision-recall curve, where 
precision equals TP\,/\,(TP+FP), 
recall equals TP\,/\,(TP+FN), 
and TP, FP, and FN correspond to the number of true-positive, false-positive, and false-negative detections, respectively. 
For reproducability, we follow the suggested metric implementations of the datasets, which approximate the area under the curve at sampling points. For the association of predicted and ground truth bounding boxes, the BEV center point distance is used for nuScenes, and the intersection-over-union ratio in 2D, BEV and 3D is used for \mbox{View-of-Delft}. For nuScenes, some additional metrics are calculated for the true-positive detections, including translation, scale, orientation, velocity, and attribute errors.

\subsection{Quantitative results}

In~Table~\ref{table:nuscenes_AP}, we show the AP results on nuScenes for classes pedestrian, cyclist, and car. We observe that the radar-only network is able to detect cars to a certain degree but it can barely detect pedestrians and cyclists, as these often do not have a single radar detection and are difficult to distinguish from the environment. The camera-only network does better across all classes and shows that with a large dataset like nuScenes, monocular 3D object detection is possible. When fusing radar and camera, we see significant improvements for all classes, demonstrating how well the combination of the semantic value from the camera and the geometric value from the radar can work together.

\begin{table}[hbt]
\caption{Experimental validation results for radar-camera fusion in BEV on nuScenes front-view: average precision metrics.}
\label{table:nuscenes_AP}
\begin{center}
\begin{tabular}{|c|r|r|r|r|}
\hline
{\bf Modality} & {\bf mAP}  & {\bf AP} {\tt ped} & {\bf AP} {\tt cyc} & {\bf AP} {\tt car} \\ \hline
Radar        & 12.5 &  3.1 &  5.4 & 29.0 \\
Camera       & 27.7 & 26.7 & 17.3 & 39.1 \\
Radar-camera & 40.6 & 36.6 & 26.4 & 58.9 \\ \hline
\end{tabular}
\end{center}
\end{table}

In~Table~\ref{table:nuscenes_TP}, we show the detailed true-positive metrics on nuScenes for each class. 
It should be noted that the radar-only metrics for pedestrian and cyclist are not meaningful due to the very low AP values for these classes. 
Comparing radar-only and camera-only performance for cars, we can see that despite the lower AP, the radar does better in translation, orientation, velocity, and nuScenes attribute estimation. This is deemed due to the direct distance and velocity measurement. The radar-camera fusion further improves all metrics, again showing the synergies of the different modalities.

\begin{table*}[hbt]
\caption{Experimental validation results for radar-camera fusion in BEV on nuScenes front-view: true-positive metrics, including average errors for translation (ATE), scale (ASE), orientation (AOE), velocity (AVE), and attribute (AAE). For details on the metrics refer to \cite{nuScenes2020}.} 
\label{table:nuscenes_TP}
\begin{center}
\begin{tabular}{|c|rrrrr|rrrrr|rrrrr|}
\hline
{\bf Modality} & \multicolumn{5}{c|}{{\bf AP} {\tt ped}} & \multicolumn{5}{c|}{{\bf AP} {\tt cyc}} & \multicolumn{5}{c|}{{\bf AP} {\tt car}} \\
& ATE & ASE & AOE & AVE & AAE & ATE & ASE & AOE & AVE & AAE & ATE & ASE & AOE & AVE & AAE \\ \hline
Radar         & 0.53 & 0.31 & 0.71 & 0.67 & 0.11 & 0.52 & 0.34 & 0.45 & 0.11 & 0.02 & 0.44 & 0.20 & 0.23 & 0.60 & 0.11 \\
Camera        & 0.77 & 0.30 & 1.40 & 0.83 & 0.73 & 0.76 & 0.28 & 1.13 & 0.52 & 0.01 & 0.64	& 0.17 & 0.36 & 1.72 & 0.31 \\
Radar-camera  & 0.49 & 0.30 & 0.88 & 0.63 & 0.27 & 0.45 & 0.28 & 0.93 & 0.29 & 0.03 & 0.31 & 0.17 & 0.14 & 0.44 & 0.23 \\ \hline
\end{tabular}
\end{center}
\end{table*}

\begin{table*}[ht]
\caption{Experimental validation results for radar-camera fusion in BEV on \mbox{View-of-Delft}: average precision metrics.}
\label{table:view-of-delft_AP}
\begin{center}
\begin{tabular}{|c|c|rrr|rrr|rrr|rrr|}
\hline
{\bf Modality} & {\bf Pretrain} & \multicolumn{3}{c|}{\bf mAP} & \multicolumn{3}{c|}{{\bf AP} {\tt ped}} & \multicolumn{3}{c|}{{\bf AP} {\tt cyc}} & \multicolumn{3}{c|}{{\bf AP} {\tt car}} \\
& & 2D & BEV & 3D & 2D & BEV & 3D & 2D & BEV & 3D & 2D & BEV & 3D \\ \hline
Radar         & & 48.5 & 49.1 & 43.7 & 34.1 & 32.0 & 29.4 & 68.3 & 66.6 & 65.4 & 43.2 & 48.7 & 36.3 \\
Camera        & & 32.1 & 14.2 & 12.1 & 24.5 & 9.8 & 9.6 & 40.5 & 14.2 & 12.5 & 31.3 & 18.5 & 14.0 \\
Radar-camera  & & 53.1 & 53.5 & 48.0 & 43.2 & 40.6 & 37.2 & 71.5 & 67.7 & 67.2 & 44.6 & 52.3 & 39.7 \\ \hline
Camera       & $\checkmark$ & 49.6 & 21.5 & 16.9 & 39.2 & 12.6 & 11.4 & 58.5 & 26.6 & 21.6 & 51.2 & 25.4 & 17.7 \\
Radar-camera & $\checkmark$ & 58.2 & 55.7 & 47.7 & 50.0 & 39.4 & 35.0 & 72.0 & 69.2 & 67.7 & 52.7 & 58.5 & 40.4 \\ \hline
\end{tabular}
\vspace{-2mm}
\end{center}
\end{table*}

For \mbox{View-of-Delft}, we show the AP results in terms of 2D, BEV, and 3D metrics in Table~\ref{table:view-of-delft_AP}. 
With the high-performance radar, the radar-only approach works quite well for all classes, and even achieves the best results for cyclists, which were difficult to detect in nuScenes. 
The camera-only network performs worse than radar-only for all classes. On the 2D image plane, the results are more competitive, but we clearly observe the difficulty of the monocular depth estimation in the BEV and 3D metrics, given the limited amount of training data in \mbox{View-of-Delft}. 
The fusion network once again outperforms the single-modality networks. 
To provide more variety in the training images used for the camera branch, we pretrain the camera-only network with nuScenes and fine-tune the camera-only and radar-camera fusion networks with \mbox{View-of-Delft}. 
For camera-only, we observe significant improvements in all metrics. The radar-camera fusion network benefits from the pretraining mostly in the 2D metrics, since the BEV and 3D metrics are dominated by the high-performance radar in \mbox{View-of-Delft}.

\subsection{Exemplary qualitative results}

\begin{figure*}[b]
\centering
\begin{subfigure}{.275\textwidth}
  \centering
  \includegraphics[width=0.95\textwidth]{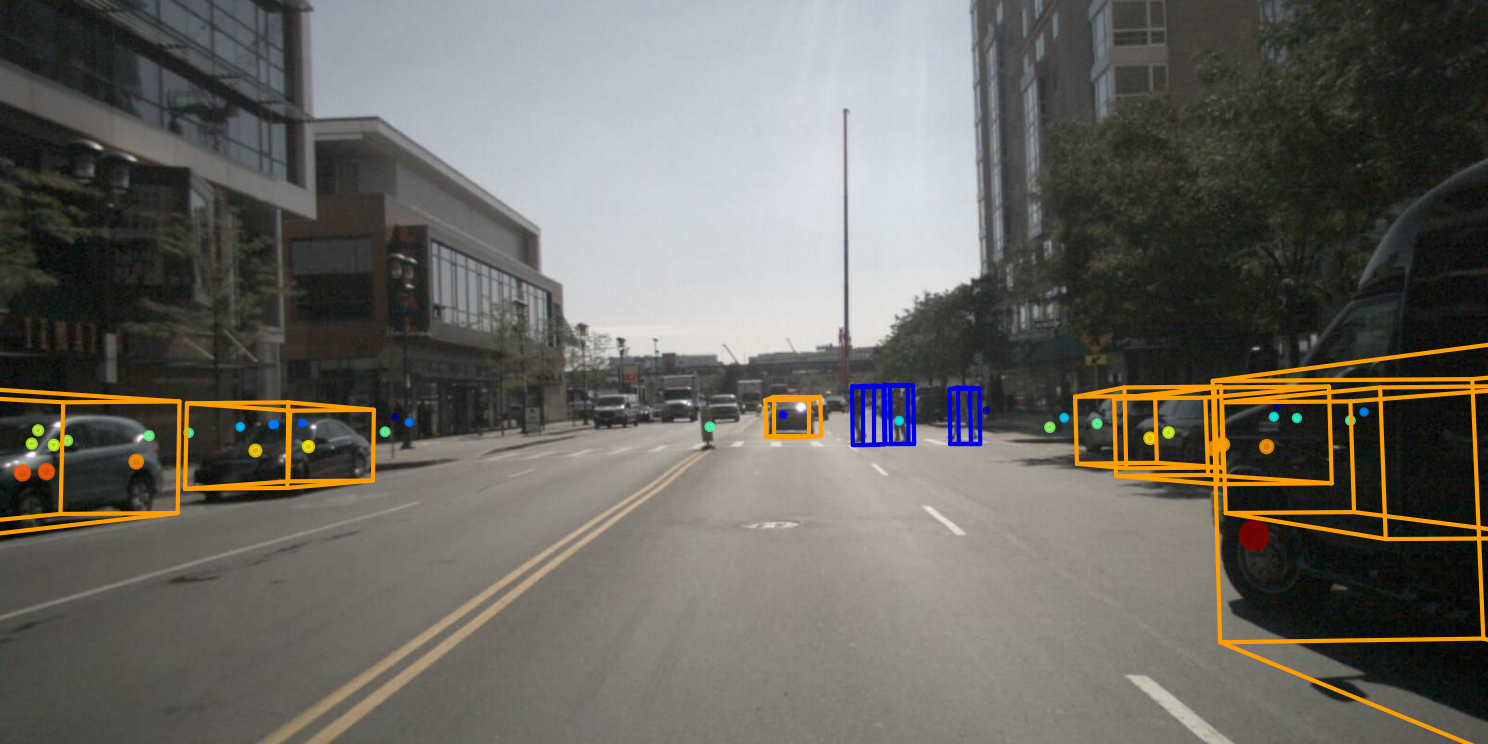}
  \includegraphics[width=0.95\textwidth]{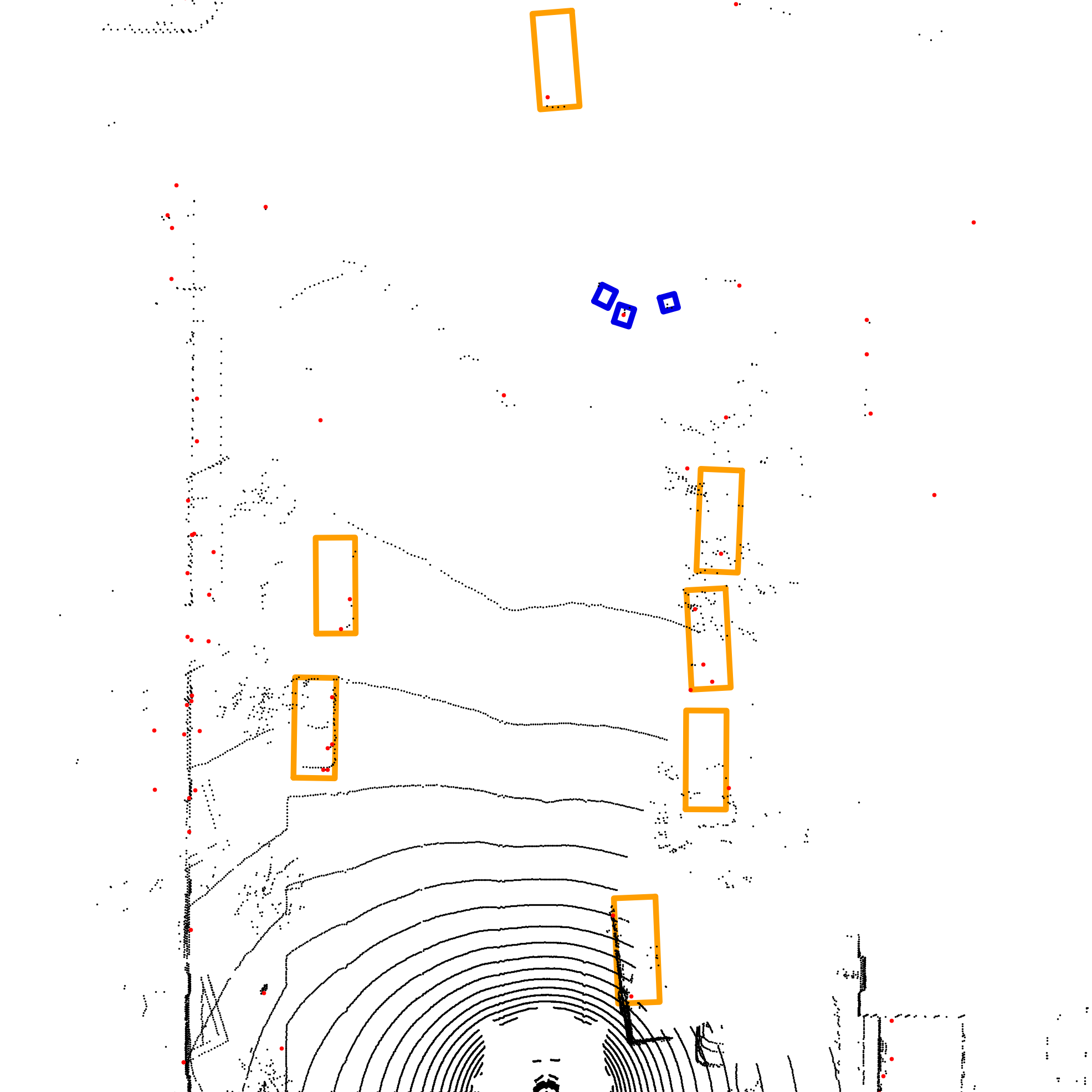}
  \caption{}
  \label{fig:sub1}
\end{subfigure}%
\begin{subfigure}{.275\textwidth}
  \centering
  \includegraphics[width=0.95\textwidth]{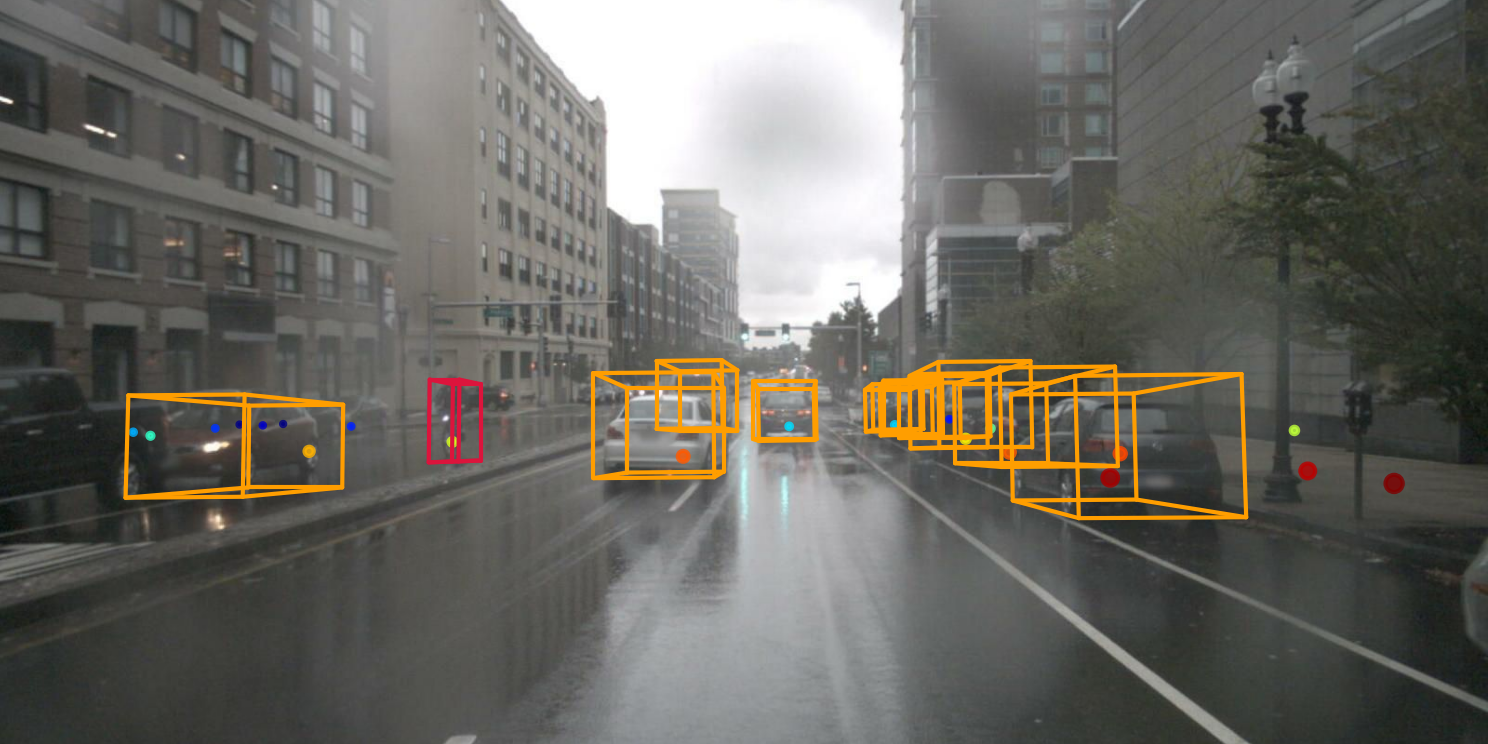}
  \includegraphics[width=0.95\textwidth]{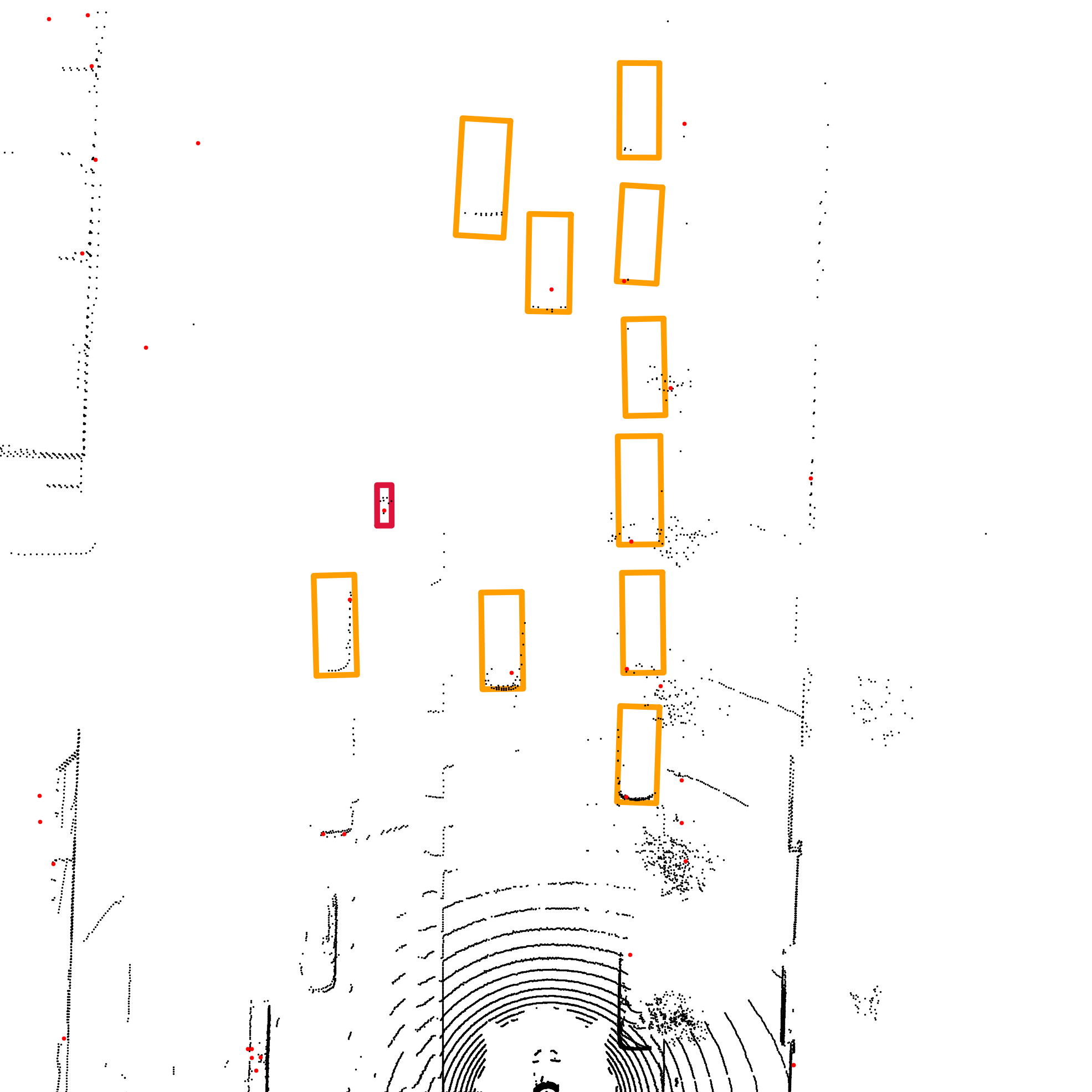}
  \caption{}
  \label{fig:sub2}
\end{subfigure}%
\begin{subfigure}{.275\textwidth}
  \centering
  \includegraphics[width=0.95\textwidth]{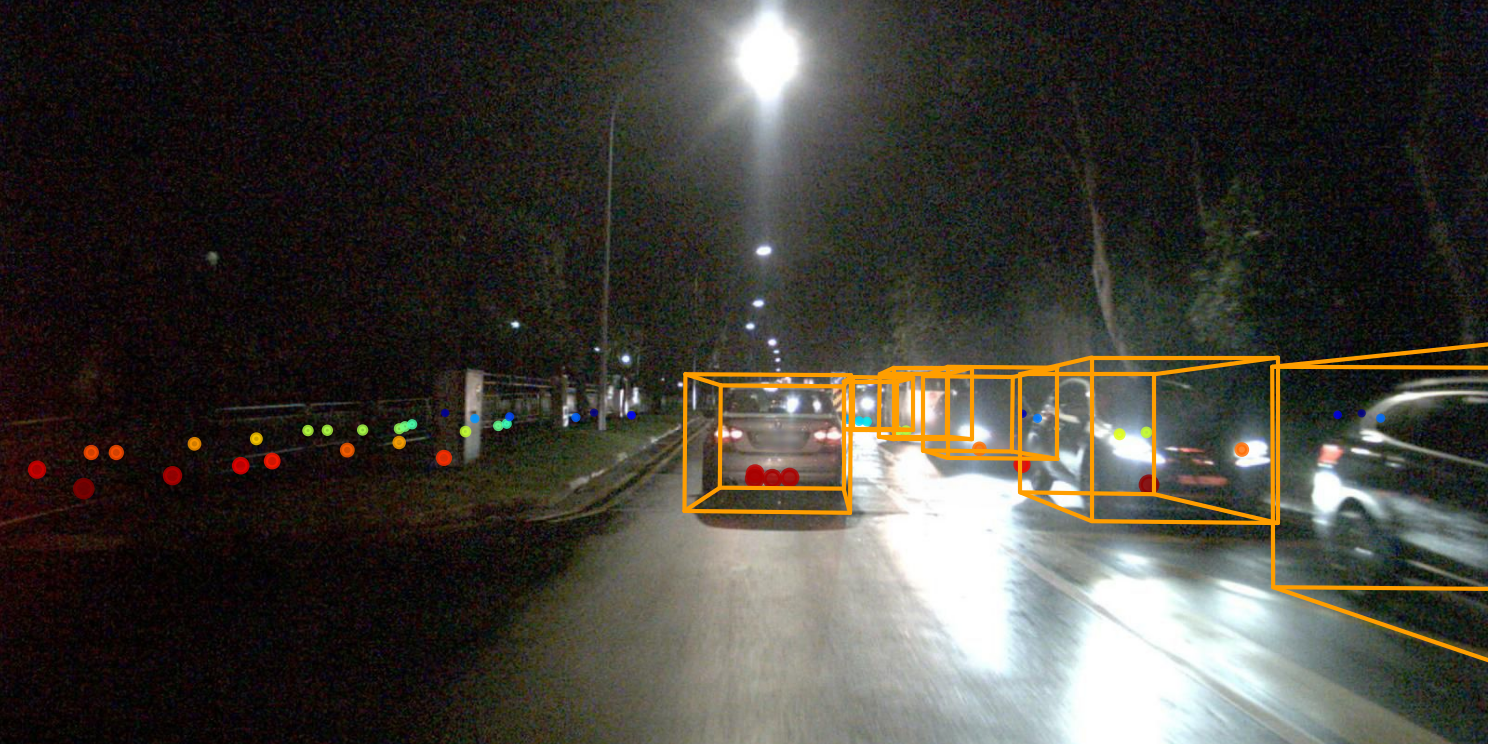}
  \includegraphics[width=0.95\textwidth]{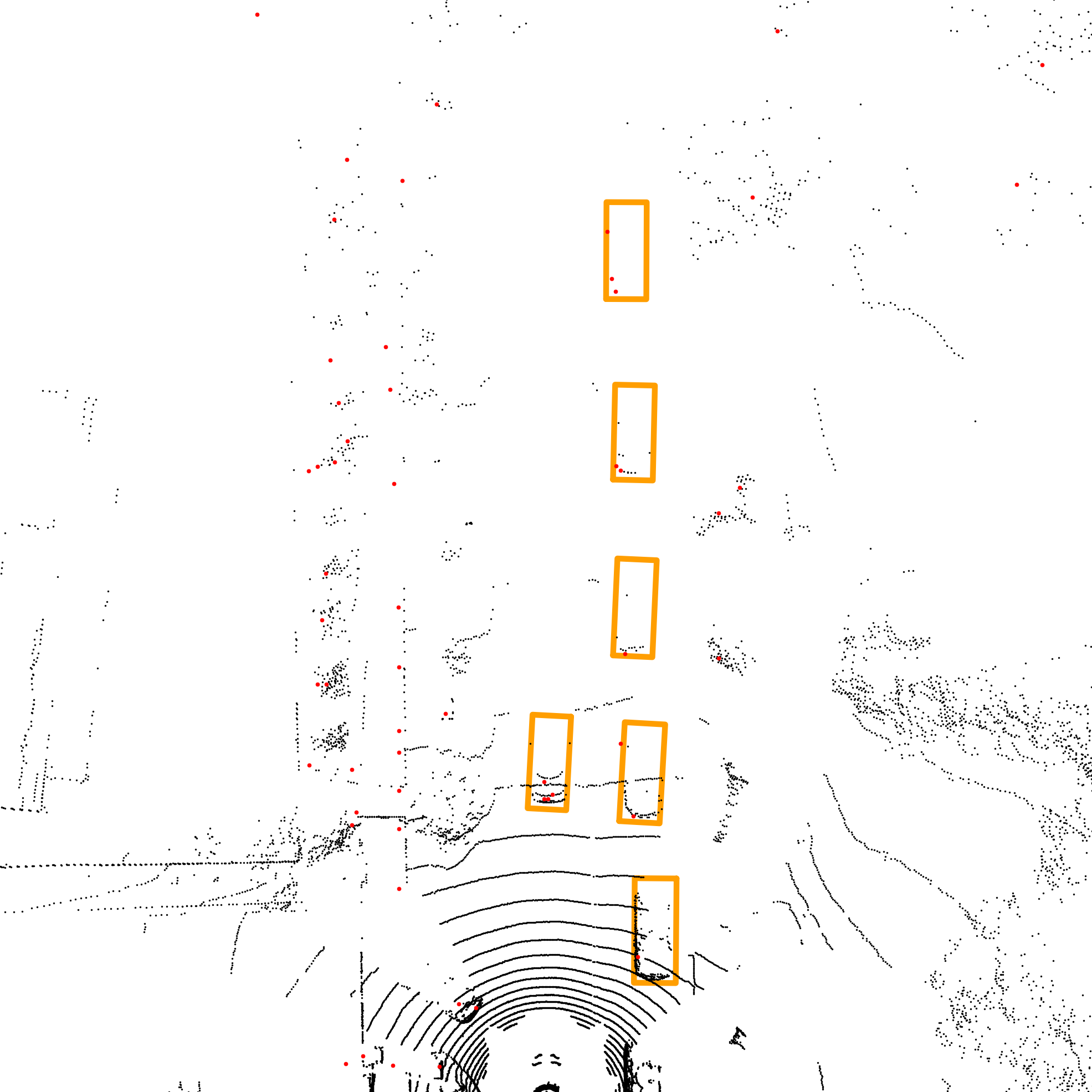}
  \caption{}
  \label{fig:sub3}
\end{subfigure}\vspace{1mm}
\begin{subfigure}{.275\textwidth}
  \centering
  \includegraphics[width=0.95\textwidth]{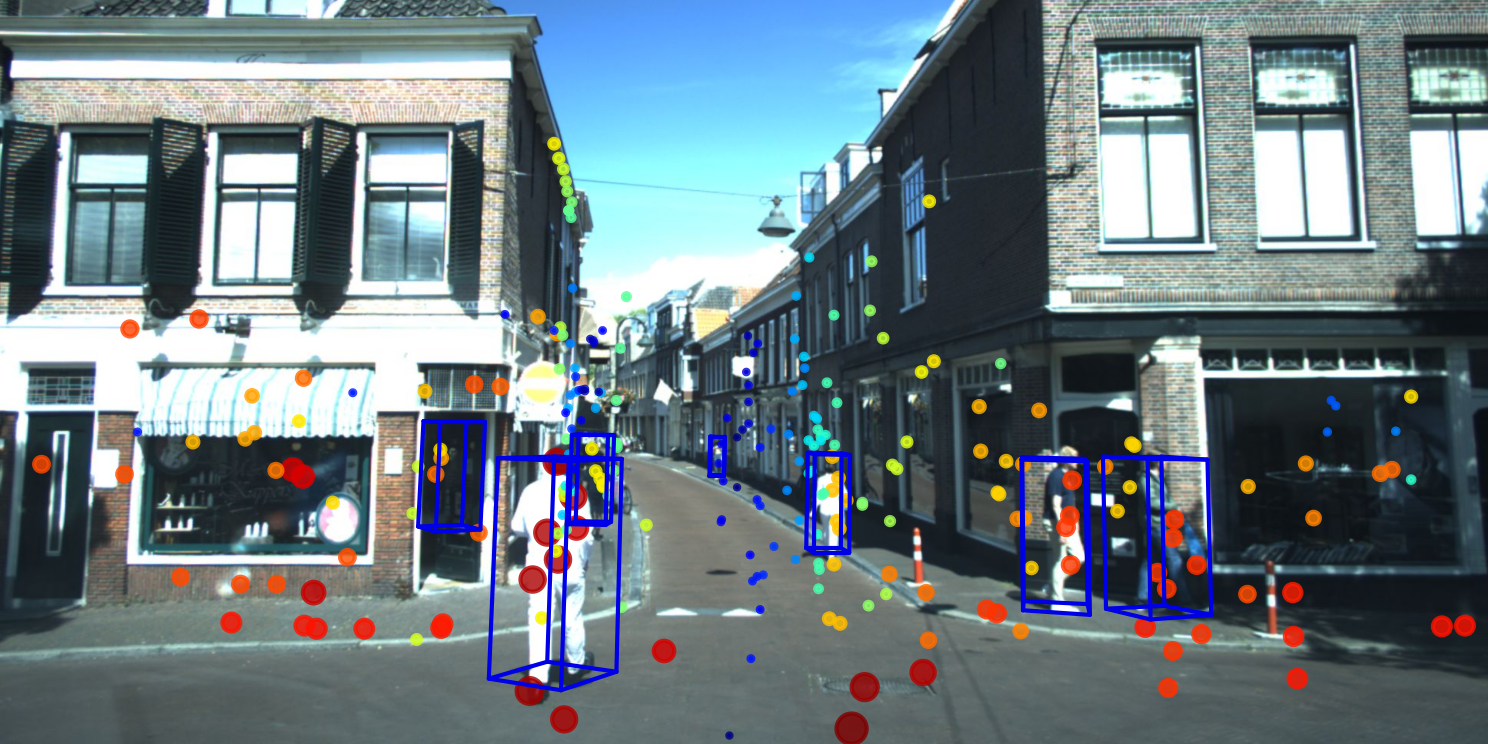}
  \includegraphics[width=0.95\textwidth]{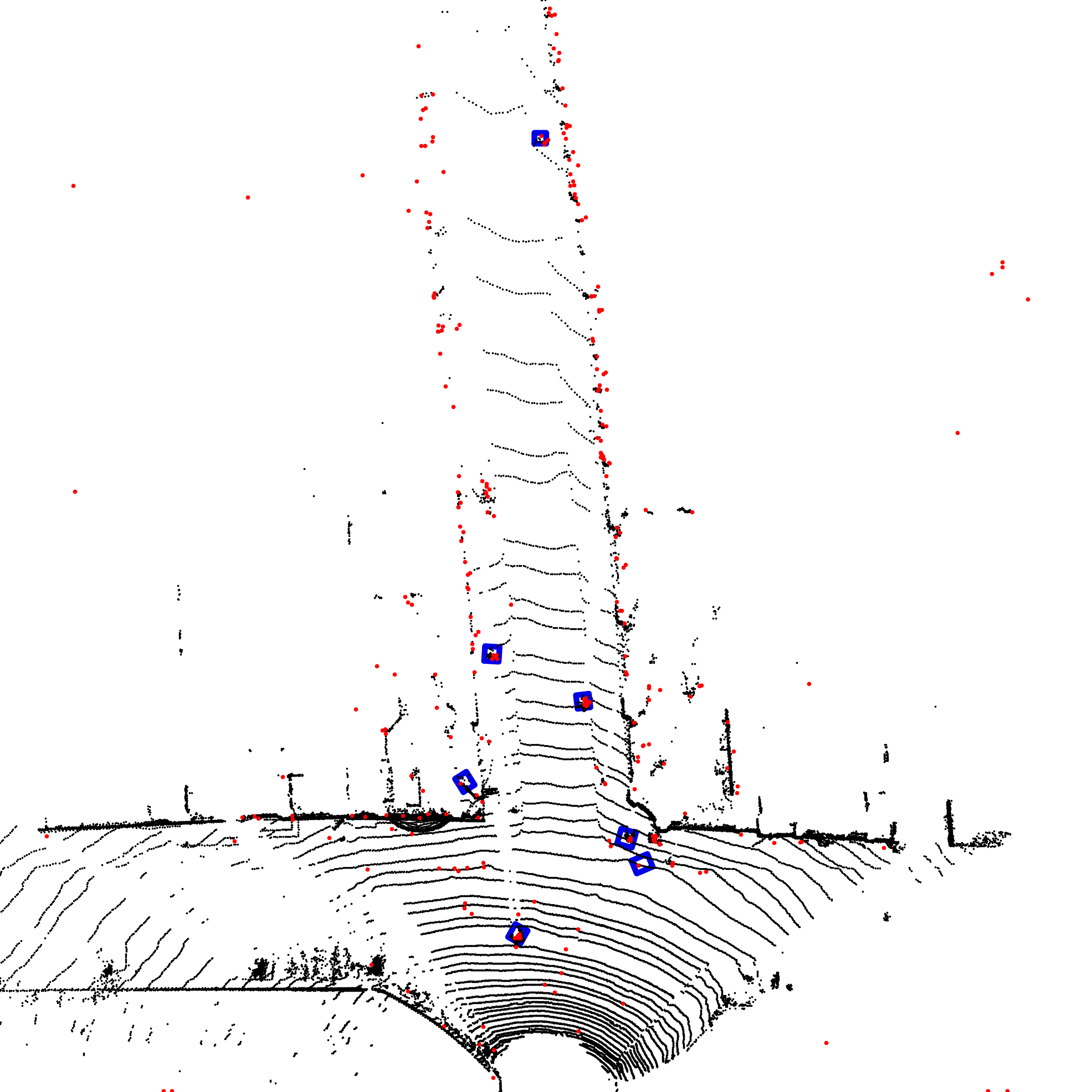}
  \caption{}
  \label{fig:sub4}
\end{subfigure}%
\begin{subfigure}{.275\textwidth}
  \centering
  \includegraphics[width=0.95\textwidth]{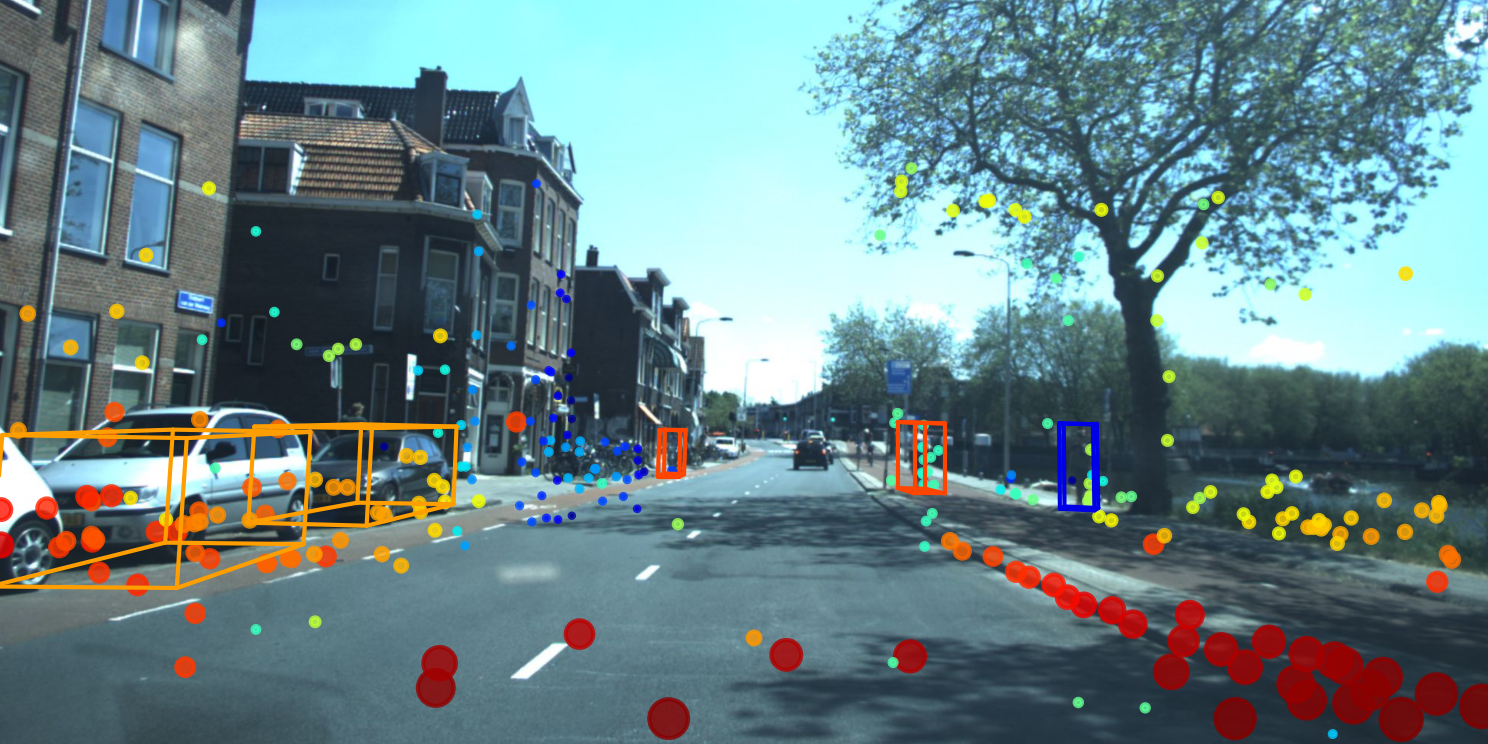}
  \includegraphics[width=0.95\textwidth]{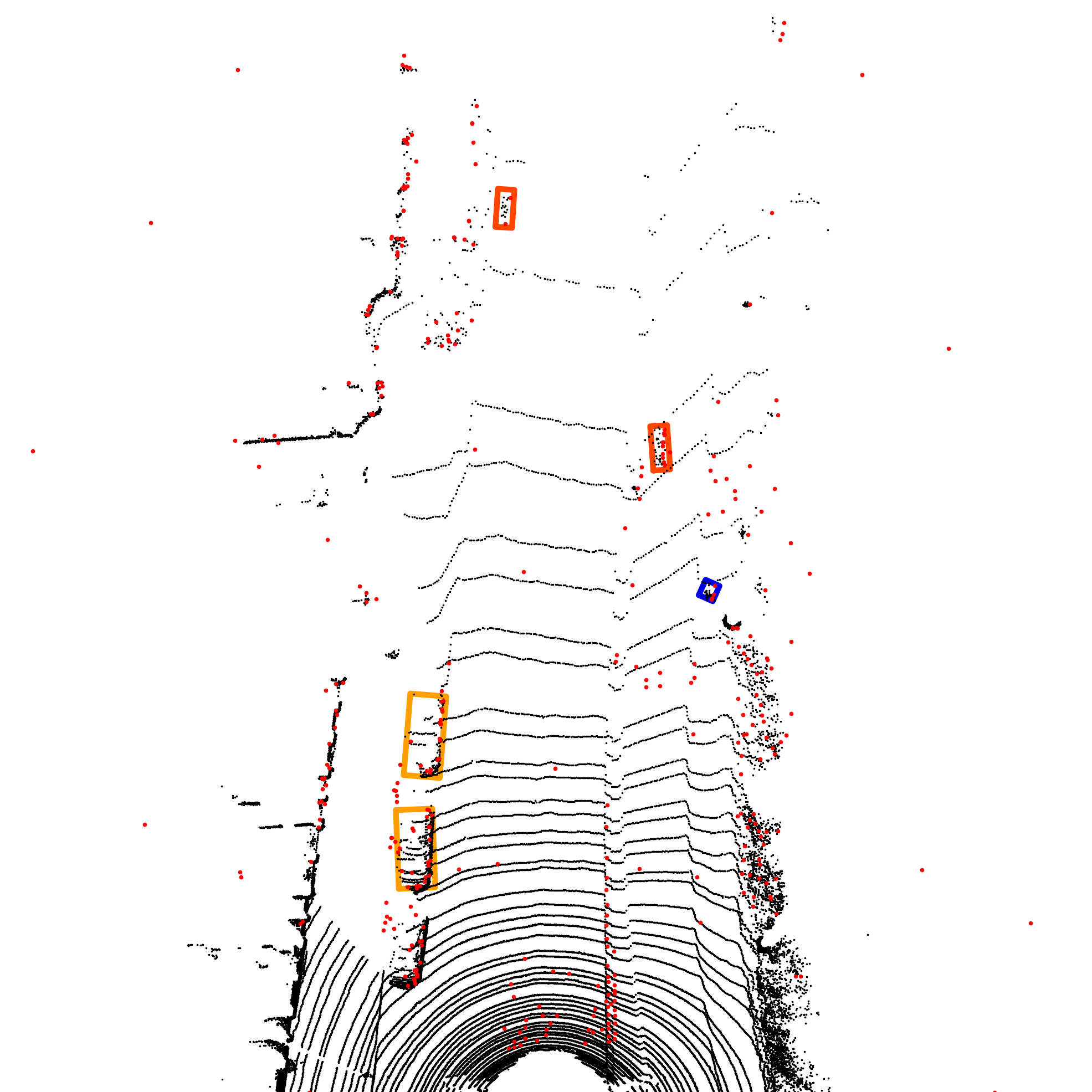}
  \caption{}
  \label{fig:sub5}
\end{subfigure}%
\begin{subfigure}{.275\textwidth}
  \centering
  \includegraphics[width=0.95\textwidth]{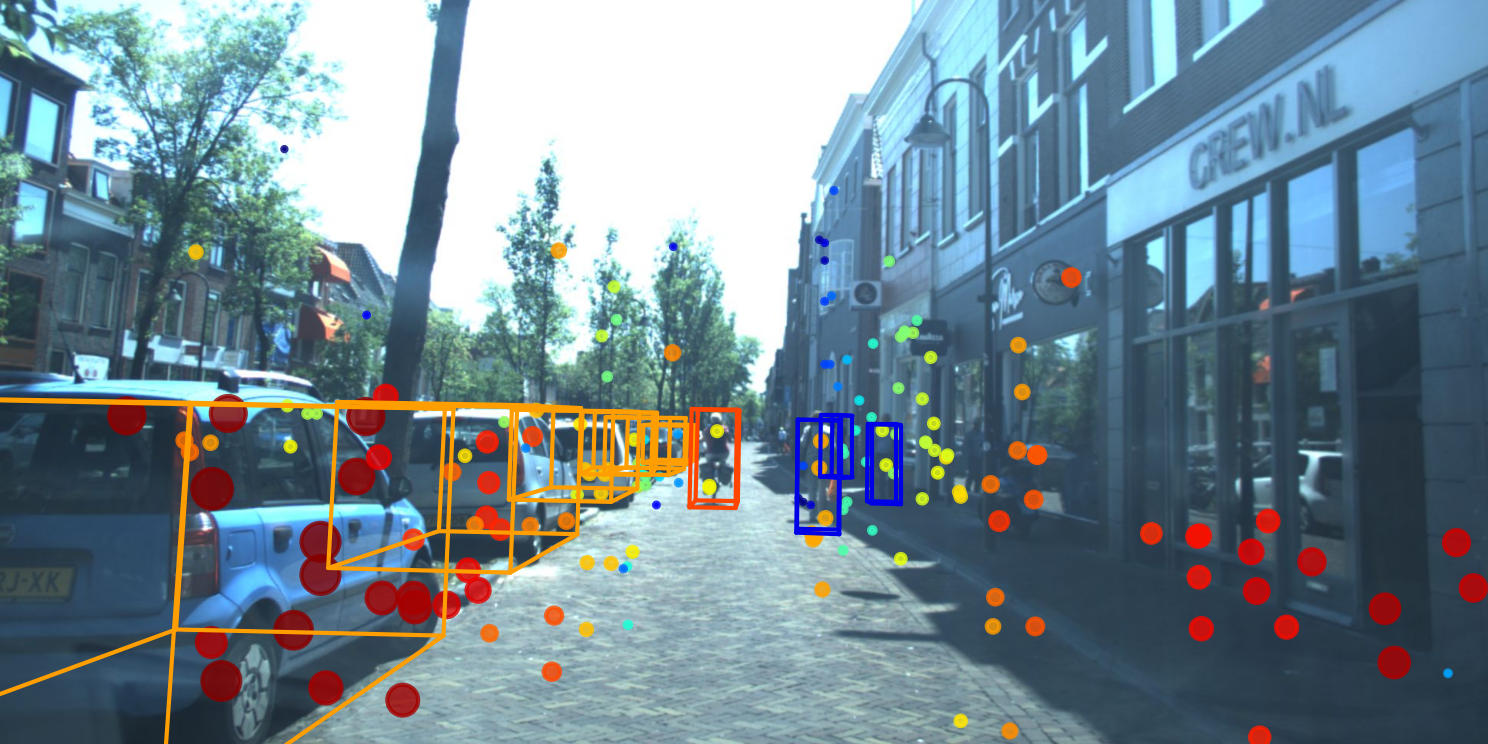}
  \includegraphics[width=0.95\textwidth]{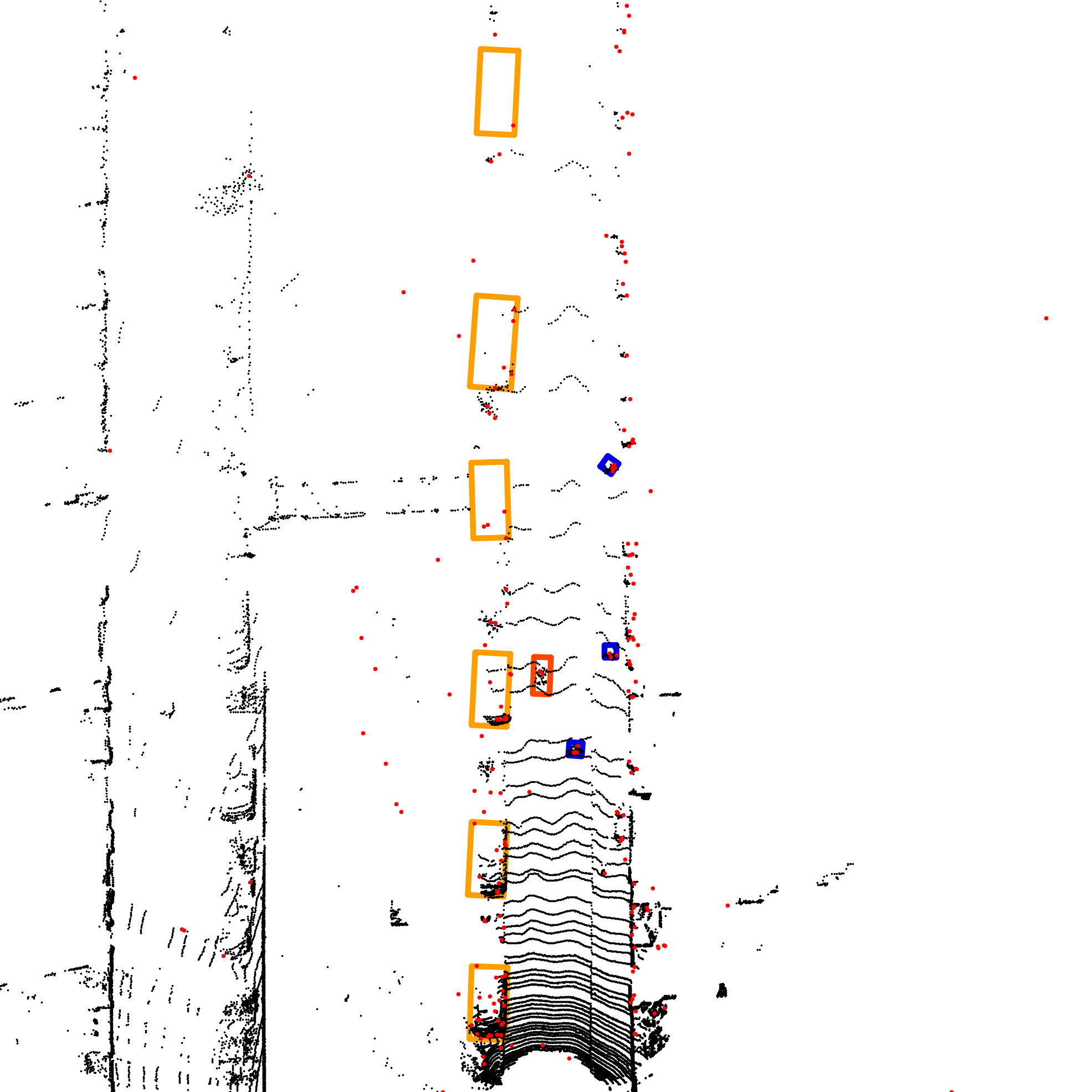}
  \caption{}
  \label{fig:sub6}
\end{subfigure}
\caption{Qualitative radar-camera fusion results from nuScenes (a)--(c) and \mbox{View-of-Delft} (d)--(f). The respective upper and lower plots show the camera image and the BEV grid, including projected radar detections and predicted bounding boxes, blue for pedestrian, red for cyclist, and orange for cars. The LiDAR point cloud in the BEV grid is shown for reference only.}
\label{fig:qualitative}
\end{figure*}

To demonstrate the performance of our best-performing radar-camera fusion models, we show some selected qualitative examples for each dataset in Figure~\ref{fig:qualitative}. In each subfigure, we show the frontal camera image with the projected radar detections and a BEV representation of the same frame with radar detections in red and the LiDAR point cloud in black. Note that the LiDAR point cloud is shown as a geometric reference only. The colored boxes are the projected 3D bounding boxes, predicted by our fusion models, where blue is used for pedestrians, red for cyclists, and orange for cars. 

The examples in \ref{fig:sub1}--\ref{fig:sub3} are from nuScenes, whereas the examples in \ref{fig:sub4}-\ref{fig:sub6} are from \mbox{View-of-Delft}. When looking at the raw data, we can observe the strengths and weaknesses of the two datasets. In particular, nuScenes offers a large variety in the images, since the scenes are recorded in two different countries, and at different daytimes and weather conditions. However, the radar point cloud is very sparse and does not contain elevation angles. In contrast, \mbox{View-of-Delft} has less variety in the images, but a much denser radar point cloud with elevation angles.

In all six frames, we can see how well our radar-camera fusion models performs. Even though we selected dense traffic scenes with many cars and pedestrians, and some cyclists, the network is able to detect and localize all the relevant objects in the scenes. Especially in the challenging situation for the camera
in \ref{fig:sub3}, we can see how the radar can help to accurately localize the cars.

\section{Conclusion}

In this paper, we have used a novel radar-camera fusion network on the BEV plane to study differences in the nuScenes and \mbox{View-of-Delft} datasets. Our results show that camera-only 3D object detection requires a large dataset with reasonable visual variability, as it is available in the nuScenes dataset. In contrast, the radar-only network profits more from the high-performance radar in \mbox{View-of-Delft} and can cope with a smaller dataset. Also, it can also classify vulnerable road users like pedestrians and cyclists.

We conclude that the full potential of radar-camera fusion could be achieved when combining the needs for radar and camera perception, with a dense point cloud and a large visual variability, respectively. Until such a dataset is publicly available, we have shown that pretraining the camera-only network with a large dataset like nuScenes can help to improve the performance of camera-only and radar-camera fusion networks in smaller datasets like \mbox{View-of-Delft}.
In future work, we want to examine more of the recently introduced radar datasets to support our findings and investigate further transfer learning possibilities. 

\section*{Acknowledgment}
This work is partly funded by the German Federal Ministry for Economic Affairs and Climate Action (BMWK) and partly financed by the European Union in the frame of NextGenerationEU within the project ”Solutions and Technologies for Automated Driving in Town” (FKZ 19A22006P).


\begin{thebibliography}{00}


\bibitem{EngelsEtal2021} F.~Engels, P.~Heidenreich, M.~Wintermantel, L.~St\"acker, M.~Al~Kadi, and~A.~Zoubir, ``Automotive radar signal processing: Research directions and practical challenges,'' IEEE J. Sel. Top. Signal Process., vol.~15, no.~4, pp.~865--878, 2021.

\bibitem{ZhouEtal2022}
Y.~Zhou, L.~Liu, H.~Zhao, M.~Lopez-Benitez, L.~Yu, and Y.~Yue, ``Towards deep radar perception for autonomous driving: Datasets, methods, and challenges,'' Sensors, vol.~22, no.~11, 2022.

\bibitem{nuScenes2020} 
H.~Caesar {\it et al.}, ``nuScenes: A multimodal dataset for autonomous driving,'' in Proc. IEEE Comput. Soc. Conf. Comput. Vis. Pattern Recognit., 2020.

\bibitem{ViewOfDelft2022} A.~Palffy, E.~Pool, S.~Baratam, J.~Kooij, and~D.~Garvilla, ``Multi-Class Road user detection with 3+1D radar in the View-of-Delft dataset,'' IEEE Robot. and Autom. Lett., vol.~7, no.~2, pp.~4961--4968, 2022.

\bibitem{BEVFusion2023}
Z.~Liu {\it et al.}, ``BEVFusion: Multi-task multi-sensor fusion with unified bird's-eye view representation,'' in Proc. IEEE Int. Conf. Robot. Autom., 2023.

\bibitem{RadarScenes2021}
O.~Schumann {\it et al.}, ``RadarScenes: A real-world radar point cloud data set for automotive applications,'' in Proc. IEEE Int. Conf. Inf. Fusion, 2021.

\bibitem{aiMotive2022}
T.~Matuszka {\it et al.}, ``aiMotive dataset: A multimodal dataset for robust autonomous driving with long-range perception,'' arXiv preprint, arXiv:2211.09445, 2022.

\bibitem{Astyx2019}
M.~Meyer and G.~Kuschk, ``Automotive radar dataset for deep learning based 3D object detection,'' in Proc. Eur. Radar Conf., 2019.

\bibitem{TJ4DRadSet2022}
L.~Zheng {\it et al.}, ``TJ4DRadSet: A 4D radar dataset for autonomous driving,'' in Proc. IEEE Int. Conf. Intell. Transport. Syst., 2022.

\bibitem{KRadar2022}
D.-H.~Paek, S.-H.~Kong, and K.~Tirta~Wijaya, ``K-Radar: 4D radar object detection for autonomous driving in various weather conditions,'' in Proc. Conf. on Neural Inf. Process. Syst., 2022.

\bibitem{Chadwick2019}
S.~Chadwick, W.~Maddetn and P.~Newman, ``Distant vehicle detection using radar and vision,'' in Int. Conf. Robot. Autom., pp.~8311–8317, 2019.


\bibitem{Fei.2020}
J.~Fei, W.~Chen, P.~Heidenreich, S.~Wirges, and C.~Stiller, ``SemanticVoxels: Sequential fusion for 3D pedestrian detection using LiDAR point cloud and semantic segmentation,'' in Proc. IEEE Int. Conf. Multisensor Fusion and Integration, 2020.

\bibitem{CenterFusion2021}
R.~Nabati and H.~Qi, ``Centerfusion: Center-based radar and camera fusion for 3d object detection,'' in Proc. IEEE Winter Conf. Appl. Comput. Vis., pp.~1527-1536, 2021.

\bibitem{LiftSplatShoot2020}
J.~Philion and S.~Fidler, ``Lift, splat, shoot: Encoding images from arbitrary camera rigs by implicitly unprojecting to 3D,'' in Proc. Eur. Conf. Comput. Vis., 2020.

\bibitem{BEVDet2021}
J.~Huang, G.~Huang, Z.~Zhu, Y.~Ye, and D.~Du, ``BEVDet: High-performance multi-camera 3D object detection in bird-eye-view,'' arXiv preprint, arXiv:2112.11790, 2021.

\bibitem{PointPillars2019}
A.~Lang, S.~Vora, H.~Caesar, L.~Zhou, J.~Yang, and O.~Beijbom, ``PointPillars: Fast encoders for object detection from point clouds,'' in Proc. IEEE Conf. Comput. Vis. Pattern Recog., 2019.

\bibitem{SwinTransform2021}
Z.~Liu {\it et al.}, ``Swin transformer: Hierarchical vision transformer using shifted windows,'' in Proc. Int. Conf. Comput. Vis., 2021.

\bibitem{CenterPoint2021}
T.~Yin, X.~Zhou, and P.~Kr\"ahenb\"uhl, ``Center-based 3D object detection and tracking,'' 
in Proc. IEEE Conf. Comput. Vis. Pattern Recog., 2021.

\bibitem{CGBS2019}
B.~Zhu, Z.~Jiang, X.~Zhou, Z.~Li, and G.~Yu. ``Class-balanced grouping and sampling for point cloud 3D object detection. arXiv preprint, arXiv:1908.09492, 2019.

\end{thebibliography}
\end{document}